\pdfoutput=1

\documentclass[11pt]{article}

\usepackage{acl}

\usepackage{float}
\usepackage{times}
\usepackage{multirow}
\usepackage{booktabs}
\usepackage{microtype}
\usepackage{latexsym}
\usepackage{booktabs}
\usepackage{courier}
\usepackage{amsmath}
\usepackage{subfig}
\usepackage{placeins}
\usepackage{tikz}
\usepackage{hyperref}

\usepackage[T1]{fontenc}

\usepackage[utf8]{inputenc}
\usepackage{graphicx}

\usepackage{microtype}

\title{SemEval-2023 Task 7: Multi-Evidence Natural Language Inference for Clinical Trial Data}

\author{\\ \textbf{Maël Jullien\textsuperscript{1}, Marco Valentino\textsuperscript{2}\textsuperscript{, 1}, Hannah Frost\textsuperscript{1}\textsuperscript{, 3}, Paul O'Regan\textsuperscript{3}}, \textbf{Donal Landers\textsuperscript{3}, Andr\'e Freitas\textsuperscript{1}\textsuperscript{, 2}}\\  Department of Computer Science, University of Manchester, United Kingdom\textsuperscript{1} \\  Idiap Research Institute, Switzerland\textsuperscript{2} \\  Digital Experimental Cancer Medicine Team, Cancer Research UK Manchester Institute\textsuperscript{3} \\ {\tt \{firstname.surname\}\tt@manchester.ac.uk} \\ {\tt \{Paul.ORegan,Donal.Landers\}\tt @digitalecmt.org}}

\begin{document}
\maketitle
\begin{abstract}
This paper describes the results of SemEval 2023 task 7 -- Multi-Evidence Natural Language Inference for Clinical Trial Data (NLI4CT) -- consisting of 2 tasks, a Natural Language Inference (NLI) task, and an evidence selection task on clinical trial data. The proposed challenges require multi-hop biomedical and numerical reasoning, which are of significant importance to the development of systems capable of large-scale interpretation and retrieval of medical evidence, to provide personalized evidence-based care.

Task 1, the entailment task, received 643 submissions from 40 participants, and Task 2, the evidence selection task, received 364 submissions from 23 participants. The tasks are challenging, with the majority of submitted systems failing to significantly outperform the majority class baseline on the entailment task, and we observe significantly better performance on the evidence selection task than on the entailment task. Increasing the number of model parameters leads to a direct increase in performance, far more significant than the effect of biomedical pre-training. Future works could explore the limitations of large models for generalization and numerical inference, and investigate methods to augment clinical datasets to allow for more rigorous testing and to facilitate fine-tuning.

We envisage that the dataset, models, and results of this task will be useful to the biomedical NLI and evidence retrieval communities. The dataset\footnote{\url{https://github.com/ai-systems/nli4ct}}, competition leaderboard\footnote{\url{https://codalab.lisn.upsaclay.fr/competitions/8937\#learn_the_details}}, and website\footnote{\url{https://sites.google.com/view/nli4ct/}} are publicly available.

\end{abstract}

\section{Introduction}

\begin{figure}[t]
\centering
\includegraphics[width=\columnwidth]{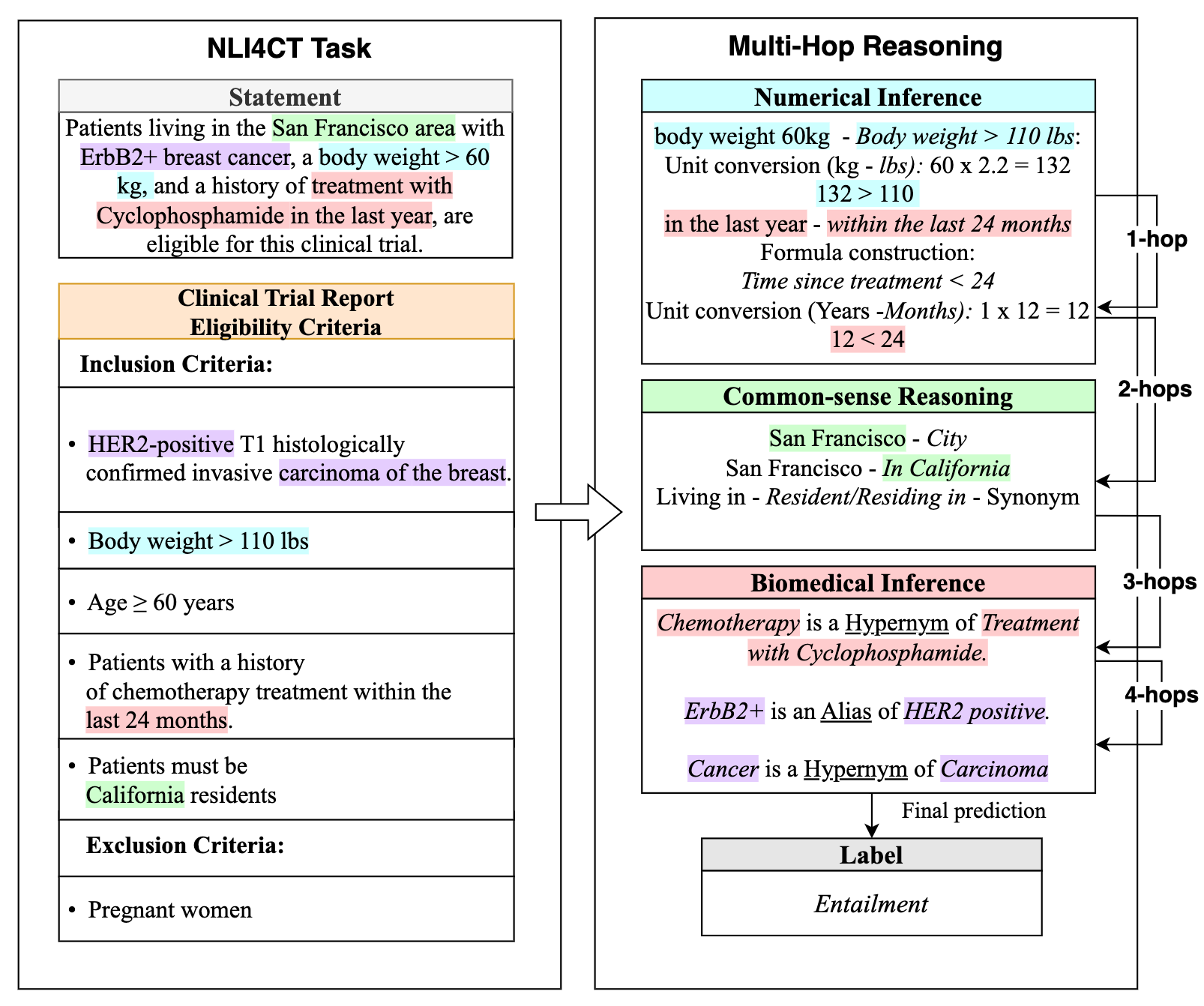}

\caption{We propose two tasks for reasoning on clinical trial data expressed in natural language. Firstly, to predict the entailment of a \textbf{Statement} and a \textbf{CTR} premise, and secondly, to extract evidence to support the label.}
\end{figure}

Clinical trials are indispensable for experimental medicine as they test the efficacy and safety of novel treatments \cite{avis2006factors}. Clinical Trial Reports (CTRs) are documents that detail the methodology and results of a trial, implemented to guide personalized and targeted interventions for patients. However, there are 400,000+ published CTRs, with an increasing number being published every year \cite{bastian2010seventy}, making it impractical to manually carry out comprehensive evaluations of all the relevant literature when designing new treatment protocols \cite{DeYoung2020EvidenceI2}.

To address this challenge, Natural Language Inference (NLI) \cite{bowman2015large,devlin-etal-2019-bert} offers a potential solution for the large-scale interpretation and retrieval of medical evidence, to support a higher level of precision and efficiency in personalized evidence-based care \cite{sutton2020overview}. 

SemEval-2023 Task 7 -- Multi-Evidence Natural Language Inference for Clinical Trial Data (NLI4CT) -- is based on the NLI4CT dataset which contains 2 tasks on breast cancer CTRs, shown in Figure 1. Firstly to determine the inference relation between a natural language statement, and a CTR. Secondly, to retrieve supporting facts from the CTR(s) to justify the predicted relation.

The inference task requires Multi-hop reasoning, that is the ability to combine information from multiple pieces of text to draw inferences \cite{jansen2018worldtree, dalvi2021explaining}. 
Previous works have shown that although multi-hop reasoning can be implemented on large-scale scientific tasks, there is a significant drop-off in performance as the number of necessary hops increases \cite{valentino2022hybrid,valentino2021unification,thayaparan2022diff,thayaparan2021explainable}. A large proportion of the NLI4CT dataset instances require the construction of inference chains in this drop-off range. 

Additionally, numerical and quantitive reasoning is required to perform inference on NLI4CT, exemplified in Figure 1. Studies have shown that transformer-based models are unable to consistently apply this type of reasoning,  instead relying on shallow heuristics for predictions \cite{DBLP:journals/corr/abs-2103-07191,DBLP:journals/corr/abs-1901-03735,DBLP:journals/corr/abs-1903-11907}. 

In the NLI4CT inference task, both the multi-hop and the numerical reasoning have the added complexity of being applied to CTRs. Studies have demonstrated that the word distribution shift from general domain corpora to biomedical corpora, such as CTRs, caused by the increased prevalence of aliases, acronyms, and biomedical terminology represents a significant detriment to model performance \cite{bio,grossman2021deep,shickel2017deep,jiang2011study,moon2015challenges,jimeno2011exploiting,pesaranghader2019deepbiowsd,jin2019deep,wu2015clinical}. 

This word distribution shift challenge is also present in the evidence selection task. Although the evidence selection task is arguably simpler than the inference task its importance cannot be understated. State-of-the-art NLI models consistently struggle to attend to relevant pieces of evidence when applied to large texts \cite{DeYoung2020EvidenceI2}. Additionally, the ability to filter out irrelevant pieces of text reduces the likelihood of distractors \cite{mishra-sachdeva-2020-need} and reduces the length of the input for inference, improving efficiency.

This paper introduces SemEval-2023 Task 7 -- Multi-Evidence Natural Language Inference for Clinical Trial Data (NLI4CT) -- for biomedical NLI and evidence extraction and presents a detailed analysis of the performance of the participating systems. We report the following conclusions;
\begin{itemize}
    \item The highest scoring system \cite{THiFLY-2023-nli4ct} @THiFLY achieved an F1 score of 0.856 and 0.853 on the entailment task and the evidence selection task respectively. 
    \item The tasks are challenging, most submissions did not significantly outperform the majority class baseline on the entailment task.
    \item On average, performance on the evidence selection task was higher than on the entailment task.
    \item Increasing the number of model parameters leads to a direct improvement in performance, far out-weighting the effect of biomedical pre-training.
\end{itemize}

\section{Related Works}

There are many existing expert-annotated resources for clinical NLI. The TREC 2021 Clinical Track \cite{soboroff2021overview} is a large-scale information retrieval task to match patient descriptions to clinical trials for which they are eligible. Evidence Inference 2.0 \cite{DeYoung2020EvidenceI2} is a Question-Answering (QA) task and span selection task, where provided with an outcome, an intervention, and a comparator intervention, systems must infer if the intervention resulted in a significant increase, significant decrease, or
produced no significant difference in the outcome measurement, compared to the comparator, and identify spans that support this inference. The MEDNLI \cite{romanov2018lessons} dataset is an entailment task to infer the entailment relation between a short piece of text extracted from medical history notes, and an annotated statement.

None of the aforementioned tasks encompass the full complexity of NLP over CTRs, that this the capability to reason over all sections CTRs and to simultaneously carry out biomedical and numerical inference. Instead choosing to focus on one specific CTR section. Additionally, these tasks often have repetitive inference chains, i.e. matching statements for eligibility, or retrieving measurements and comparing them. In contrast, NLI4CT instances cover all CTR sections and contain minimal repetition in inference chains, as there is no set template for statements.

Currently, Large Language Models (LLM) achieve the best results for clinical NLI \cite{gu2021domain, DeYoung2020EvidenceI2}. However, they suffer from a plethora of issues. LLMs demonstrate poor performance on quantitative reasoning and numerical operations within NLI \cite{DBLP:journals/corr/abs-1901-03735,DBLP:journals/corr/abs-1903-11907}. Additionally, there is a notable drop in performance for LLMs pre-trained on general domain data when applied to biomedical tasks \cite{bio}, partially aggravated by a lack of well-annotated clinical data \cite{kelly2019key}. NLI4CT is designed to assist in the development and benchmarking of models for clinical NLI.

\section{Task Description}

NLI4CT contains two tasks, Task 1, textual entailment, and Task 2, evidence selection. Each instance in NLI4CT contains a CTR premise and a statement. Premises contain 5-500 tokens, describing either the results, eligibility criteria, intervention, or adverse event of a trial, and the statements are sentences with a length of 10-35 tokens (see example in Figure 1), which make one or more claims about the premise. On average 7.74/21.67 facts within the premise are labeled as evidence. There are two types of instances in NLI4CT; single instances where the statement makes a claim about one CTR, and comparison instances, where the statement makes claims comparing and contrasting two CTRs. To summarize:

\paragraph{Task 1} Classify the inference relation between a CTR premise and a statement, as either an entailment or a contradiction, as shown in Figure 1.

\paragraph{Task 2} Output a subset of facts from the CTR premise, necessary to justify the class predicted in Task 1.

\section{Dataset}\label{Dataset}

The premises in NLI4CT are obtained from 1000 publicly available English language Breast cancer CTRs published on \href{https://clinicaltrials.gov/ct2/home}{ClinicalTrials.gov}. This data is maintained by the U.S. National Library of Medicine and is subject to the HIPAA Privacy Rule. The CTRs are split into 4 sections:

\begin{itemize}
    \item \textbf{Eligibility criteria} - A set of conditions patients must meet to participate in the trial.
    \item \textbf{Intervention} - Detailed description of the type, dosage, frequency, and duration of treatments being studied.
    \item \textbf{Results} - Reports the results of the patient cohorts in the trial with respect to a given outcome measurement.
    \item \textbf{Adverse Events} - Reports the (serious) signs and symptoms observed in patients during the clinical trial.
\end{itemize}

A group of domain experts, including clinical trial organizers from a major cancer research center, took part in the annotation task. The annotators were given two CTR premises to generate an entailment statement. This is a short text that makes an objectively true claim about the contents of the premise. Annotators could choose to write a statement about one, or both premises. Non-trivial statements typically involve summarization, comparison, negation, relation, inclusion, superlatives, aggregation, or rephrasing, and require understanding multiple rows of the premise. Then the annotators select a subset of facts from the premise(s) that support the claims in the statement.

Then a negative rewriting technique \cite{tbf} was applied, modifying the previously produced entailment statement to contain objectively false claims while retaining the original sentence structure and length. This technique is used to reduce the likelihood of stylistic or linguistic patterns pertaining to either entailment or contradictory statements. Annotators then extract a subset of facts from the premise that contradict the claims in the false statement,

The resulting dataset includes 2400 annotated statements with labels, premises, and evidence. The dataset was split 70/20/10 train/test/dev. The two classes and 
 four sections are evenly distributed throughout the dataset and its splits.

\section{Evaluation}
The same strategy is adopted for the evaluation of the results of both Tasks. Task 1, the textual entailment task, is a binary classification task, so performance is measured using Precision, Recall, and Macro F1-score, comparing predicted labels against the gold labels. We also frame Task 2, the evidence selection task, as a binary classification task, classifying each fact in the premise as either relevant evidence, or irrelevant, we compare the predicted labels against the gold labels and compute the Precision, Recall, and Macro F1-score.

\section{Architectural Paradigms}
\begin{table}[t]
    \centering
    \small
    \begin{tabular}{@{}p{4.5cm}p{2cm}cc@{}}
    \toprule
         \textbf{Technique/Model Type}  &  \textbf{Submissions \#}
\\
         \midrule
         Generative LLMs &  {\hfill}8
 \\
         \midrule
         Discriminative LLMs & {\hfill}16
\\
         \midrule
         Ontology-based &{\hfill}1
 \\
          \midrule
         Semantic rule-based &  {\hfill}1
          \\
                  \midrule
         Biomedical Pre-training &  {\hfill}12
          \\
    \bottomrule
    \end{tabular}
    \caption{Summary of the techniques and models implemented in the submissions}
    \label{tech}
\end{table}

We observe 5 different categories of approaches described in the system papers, recorded in Table 1. Generative language models are designed to learn the joint probability distribution of P(X,Y), where X is the input text, such as the statements or CTR premises, and Y is a probability output by a classification layer or a generated label from a decode-only transformer. Conversely, discriminative language models encode the conditional probability P(Y|X), designed to encode the decision boundary between different classes.  Biomedical pre-training refers to the  technique of training a model on a large, unlabeled biomedical dataset, such as scientific articles or patient health records. This is used to encode general features and patterns within a domain, before fine-tuning a specific task. Semantic rule-based models perform inference based on a set of human-defined asserted facts or axioms. Ontologies capture the categories, properties, and relations between the concepts of a particular domain. Ontology-based models extract entities from the input text and map them to nodes within the ontology to enrich the inputs with domain knowledge.

\subsection{Transformers}

\paragraph{The majority of submitted systems leverage discriminative transformers-based models.} As shown in Table 1, 16 participants integrated discriminative transformers-based models \cite{DBLP:journals/corr/VaswaniSPUJGKP17} into their submitted systems. Generally, a task-specific output layer is appended to the pre-trained layers and fine-tuned on the training to output the probability of a statement being entailed, or a piece of evidence being relevant. Alternatively, 8/21 participants submitted systems based on generative models, as seen in Table 1. These models are either appended with a task-specific output layer and fine-tuned to output a probability or directly output entailment/contradiction or relevant/irrelevant labels.

\subsection{Biomedical Pre-training}

\paragraph{The majority of participants leverage Biomedical pre-training in their systems.} As previously described LLMs trained on general domain corpora generalize poorly on biomedical corpora \cite{bio}. Therefore many participants choose to apply models that are pre-trained on biomedical texts, such as datasets of scientific articles and patient health records.

\begin{table*}[h!]
    \centering
    \tiny
    \centerline{
    \begin{tabular}{@{}p{2cm}p{4cm}p{1cm}p{1cm}p{2.3cm}p{0.5cm}p{0.5cm}p{0.5cm}p{0.5cm}p{0.5cm}p{0.5cm}cccccccccc@{}}
    \toprule
         \textbf{Work @Team name}  &\textbf{Approach}  &  \textbf{Generative/ \newline Discriminative} &\textbf{Retrieval type}&\textbf{Pre-training Datasets}&\multicolumn{3}{c}{\underline{\textbf{Task 1}}}&\multicolumn{3}{c}{\underline{\textbf{Task 2}}}
\\
&&&&&F1&Precision&Recall&F1&Precision&Recall\\
\midrule
\raggedright{\cite{THiFLY-2023-nli4ct} @THiFLY} & MGNet, BiLSTM and SciFive model ensembling & G + D & Post & \raggedright{PubMed Abstract, PMC} & 0.856 &0.856  &0.856 & 0.853	&0.811	&0.898
          \\
          \midrule
\raggedright{\cite{Saama-2023-nli4ct} @Saama AI Research }& Instruction-finetuned LLMs, Flan-T5 & G + D & - & - & 0.834&	0.768	&0.912 &  - & - &  -
          \\
          \midrule
\raggedright{\cite{Sebis-2023-nli4ct} @Sebis} & \raggedright{Ensemble of a pipeline and joint system based on DeBERTa-v3}&  D & Pre & - & 0.798	&0.777	&0.820 & 0.818	&0.772	&0.868
          \\
           \midrule
\raggedright{\cite{knowcomp-2023-nli4ct} @KnowComp }& \raggedright{DeBERTa-v3-large.
}&  D & - & - &0.764	&0.757	&0.772&  - & - &  -
          \\
\midrule
\raggedright{\cite{NCUEE-2023-nli4ct} @NCUEE-NLP} & Soft voting ensemble mechanism based on BioLink/BioBERT
&  D & Pre & \raggedright{ MultiNLI, MedNLI, and SNLI }&0.709	&0.668&	0.756&0.794&	0.803&	0.786
          \\
\midrule
\raggedright{\cite{Clemson-2023-nli4ct} @Clemson NLP }& GatorTron-BERT 
&  D & Pre & \raggedright{UFHS notes, MIMIC-III, WikiText, PMC, and extracted CTRs}&0.705	&0.654	&0.764&0.806	&0.802	&0.811
          \\
\midrule
\raggedright{\cite{I2R-2023-nli4ct} @I$^2$R }& Evidence level inferences with T5
&  G + D & Pre & - &0.701	&0.550	&0.968&0.802	&0.797	&0.807
          \\
\midrule
\raggedright{\cite{MDC-2023-nli4ct} @MDC }& PubMedBERT for evidence retrieval, and BioLinkBERT classifies entailment.
&  D & Pre & \raggedright{PubMed abstracts, PMC}&0.695	&0.668	&0.724&0.804&	0.814	&0.795
          \\
\midrule
\raggedright{\cite{HW-TSC-2023-nli4ct} @HW-TSC }& Zero-shot ChatGPT for entailment and DeBERTaV3 for retrieval.
&  G + D & Post & - &0.679	&0.592	&0.796&0.842	&0.816	&0.871
          \\
\midrule
\raggedright{\cite{BpHigh-2023-nli4ct} @BpHigh }& \raggedright{Few-shot GPT-3.5 Davinci}
&   G & - & - &0.679	&0.523	&0.968&  - & - &  -
          \\
\midrule
\raggedright{\cite{YNU-2023-nli4ct} @YNU-HPCC }& \raggedright{BioBERT, supervised contrastive learning, and back translation. }
&  D & - & PubMed Abstracts, PMC & 0.679	&0.621&	0.748&  - & - &  -
          \\
\midrule
\raggedright{\cite{kefah-2023-nli4ct} @JUST-KM} & \raggedright{Role-based Double Roberta-Large}
&  D & - & - &0.670	&0.529	&0.912&  - & - &  -
          \\
\midrule
\raggedright{\cite{SSN-2023-nli4ct} @SSNSheerinKavitha} & \raggedright{Semantic Rule based Clinical Data Analysis, TF-IDF, and BM25}
&  - & Post & - & 0.667&	0.500&	1.00&0.572	&0.542	&0.606
          \\

\midrule
\raggedright{\cite{INF-2023-nli4ct} @INF-UFRGS} & \raggedright{EvidenceSCL using a modified PairSCL model and pre-trained Biomed RoBERTa checkpoints.}
&  D & Pre & Semantic Scholar corpus  & 0.666	&0.500	&0.996&0.681&	0.615	&0.764
          \\
\midrule
\raggedright{\cite{Stanford-2023-nli4ct} @Stanford MLab} & \raggedright{Bio+Clinical/Distil/Bio Discharge Summary BERT, and ELECTRA Small ensemble}
&  D & - & MIMIC-III, PubMed Abstracts, PMC  & 0.662	&0.575	&0.780&  - & - &  -
          \\
\midrule
\raggedright{\cite{lasigeBioTM-2023-nli4ct} @lasigeBioTM }& \raggedright{Biomedical Ontology annotations, using Scispacy}
&  - & - &  -   & 0.661	&0.511&	0.936&  - & - &  -
          \\
\midrule
\raggedright{\cite{Bf3R-2023-nli4ct} @Bf3R }& \raggedright{Sentence-based BERT similarity model pre-trained on ClinicalBERT embeddings.}
&  D & Post & MIMIC III  & 0.640	&0.497&	0.900&0.671	&0.583&	0.789
          \\
\midrule
\raggedright{\cite{FII-2023-nli4ct} @FII SMART }& \raggedright{BioBERT model and
a CNN model}
&  D & - &  PubMed Abstracts, PMC & 0.596&	0.582&	0.612&  - & - &  -
          \\
          \midrule
\raggedright{\cite{FMI-2023-nli4ct} @FMI-SU }& \raggedright{Contextual Data Augmentation to fine-tune BioM-BERT-Large}
&   D & - &  PubMed Abstracts, PMC, EN Wiki + Books &-&-&-& 0.827&	0.779&	0.881
          \\
\midrule
\raggedright{\cite{CPIC-2023-nli4ct} @CPIC }& \raggedright{Ensembled  GPT-2 models with different parameter sizes and random seeds.}
&  G + D & - &  - &-&-&-& 0.810	&0.789&	0.833
          \\
          
          \midrule
\raggedright{\cite{ITTC-2023-nli4ct} @ITTC }& \raggedright{BM25 and Word Mover
Distance}
&   - & - & - &-&-&-&0.719&0.579&	0.948
          \\
    \bottomrule
    \end{tabular}}
    \caption{Summary of the techniques and models implemented in the leaderboard submissions. (G) Generative model, (D) Discriminative model, (Post) Evidence retrieved after entailment, (Pre) Evidence retrieved before entailment.}
    \label{methods}
\end{table*}

\section{Results and Discussion}

\begin{figure}[t]
\centering
\includegraphics[width=\columnwidth]{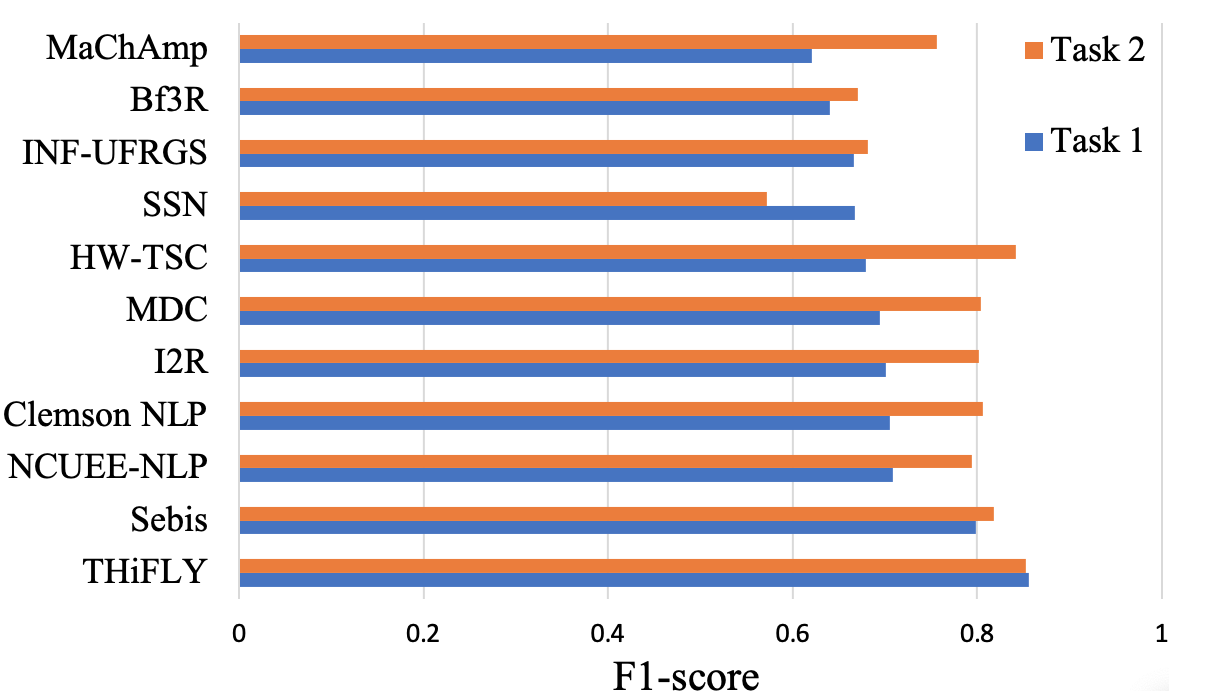}

\caption{Graph comparing system F1 scores across the entailment task and the evidence selection task}
\end{figure}
During the 21-day evaluation period (January 10$^{th}$-31$^{st}$, 2023), 40 participants submitted a total of 643 submissions for the entailment task, and 23 participants submitted a total of 364 submissions for the evidence selection task. In total 21 participants submit system papers. Submissions for which a system paper was not provided are omitted from the tables and discussion. 

\paragraph{The majority of systems fail to significantly outperform the majority-class baseline on the entailment task.}
Table 2 shows the F1 score, Recall, and Precision for Task 1. The collected results indicate that these tasks are challenging, with the majority of systems failing to achieve significantly above the majority-class baseline (0.667 F1) results on the entailment task. In particular, we observe several systems reporting 0.9-0.95 Recall, and 0.5-0.55 Precision, indicating the systems were almost exclusively predicting the "entailment" class. All systems with submitted papers significantly outperform the random baseline (0.5 F1).

\paragraph{The top-performing systems achieve significant gain across both tasks.} \citet{THiFLY-2023-nli4ct} @THiFLY, \citet{Saama-2023-nli4ct} @Saama AI Research, \citet{Sebis-2023-nli4ct} @Sebis and \citet{knowcomp-2023-nli4ct} @KnowComp achieve significantly above 0.7 F1 on the entailment task. \citet{THiFLY-2023-nli4ct} @THiFLY, \citet{HW-TSC-2023-nli4ct} @HW-TSC, \citet{FMI-2023-nli4ct} @FMI-SU, \citet{Sebis-2023-nli4ct} @Sebis, \citet{CPIC-2023-nli4ct} @CPIC, \citet{Clemson-2023-nli4ct} @Clemson NLP, \citet{MDC-2023-nli4ct} @MDC and \citet{I2R-2023-nli4ct} @@I$^2$R surpass 0.8 F1 on the evidence selection task.

\paragraph{The entailment task is more challenging than the evidence selection task.}
Table 2 shows the F1 score, Recall, and Precision for the evidence selection task. On average systems report a +0.07 higher F1 score on the evidence selection task than on the entailment task, shown in Figure 2. This result was expected as the evidence selection task does not require systems to learn complex decision boundaries between the classes or to perform numerical inference.

\paragraph{Submitted systems report higher Recall than Precision.}
On the evidence selection task, the vast majority of systems record a higher Recall than Precision, with an average difference of +0.055 higher Recall, this disparity is increasingly important with the top 5 systems, with an average difference of +0.077. A potential cause for the disparity between Precision and Recall results is statements such as "Patients with liver disease are eligible for the primary trial" where the full eligibility criteria must be returned, to provide evidence that there are no conditions against liver disease. This incentivizes systems to retrieve a large proportion of the premise, and perhaps more importantly to intentionally retrieve pieces of text that are not relevant to entities contained in the statement (liver disease). However, we hypothesize that the cost of incorrectly labeling relevant information as irrelevant is much more significant than the cost of including distracting information. This is because the entailment of a statement is often dependent on a single line of a premise. Therefore maximizing Recall, even at the cost of Precision may significantly improve evidence completeness.

\subsection{Foundational Model Architectures}

\paragraph{Generative models outperformed discriminative models on the entailment task.} As shown in Table 2 the top 2 systems on the entailment task are based on generative models, specifically 2 variants of the T5 model \cite{raffel2020exploring}, SciFive \cite{DBLP:journals/corr/abs-2106-03598} and Flan-T5 \cite{https://doi.org/10.48550/arxiv.2210.11416}. Both of these models significantly outperform the next best system with +0.058 and +0.036 F1 respectively. It should be noted that SciFive is implemented in \citet{THiFLY-2023-nli4ct} @THiFLY, as part of an ensemble with Multi-granularity Inference Networks and BiLSTMs, and therefore the system results cannot be solely attributed to the generative components.

\paragraph{DeBERTa-v3 outperforms other discriminative transformer-based models on both tasks.} DeBERTa-v3-based systems consistently outperform systems that apply discriminative models, on both tasks. This is also observed across a range of different systems settings \cite{Sebis-2023-nli4ct, HW-TSC-2023-nli4ct, knowcomp-2023-nli4ct}. DeBERTa-v3 remains competitive with the top generative approaches.

\paragraph{Increase in model size is correlated with an increase in performance.}  An increase in model size, as in models with a higher number of parameters, is strongly correlated with better performance on both Tasks. The top 5 systems in both tasks are exclusively composed of Mega Language Models (MLM) such as T5 and DeBERTa-v3-large. Additionally,  
\citet{Sebis-2023-nli4ct, Saama-2023-nli4ct} and \citet{knowcomp-2023-nli4ct} all report MLMs significantly outperforming comparatively smaller models within their individual systems.

\subsection{Rule-based systems}

\paragraph{Rule-based approaches are less competitive than MLMs.} \citet{lasigeBioTM-2023-nli4ct} @lasigeBioTM experiments with a hybrid system, using the \textit{en\_core\_sci\_lg} spaCy pipeline to extract entities from CTR premises and retrieving their ancestors from biomedical ontologies, then computing the shortest dependency path between entities, assisted with Counts and Measurements Rules to process numerical values. The premise is then combined with the premise and classified using cosine similarity. \citet{SSN-2023-nli4ct} @SSNSheerinKavitha applies a semantic rule-based system consisting of a Negation equivalence rule, Double negation rule, Deductive reasoning rule, and a Condition-based equivalence rule. Classification is obtained using TF-IDF vectors and RBF-Kernel distance similarity, and evidence is selected using BM25. As seen in Table 2, these systems are not competitive with the top-performing MLMs, however, if this disparity could be corrected, symbolic models inherently offer a higher level of transparency and interpretability than current neural models.

\subsection{Data augmentation}

\paragraph{Data-augmentation does not result in a significant performance increase.} \citet{INF-2023-nli4ct} investigates transfer learning opportunities by adding a neutral class to NLI4CT, and merging it with MultiNLI \cite{williams-etal-2018-broad} and MedNLI \cite{DBLP:journals/corr/abs-1808-06752} to train their system. \citet{FMI-2023-nli4ct} @FMI-SU annotates premise facts with structural context information, attaching trial names, cohort numbers, and parent subsection headings. They observed that the trial name does not improve performance, in some cases even adding noise, but showed some improvement with cohort and subsection annotations. \citet{Clemson-2023-nli4ct} @Clemson NLP compile an additional 9000 CTRs, and train a GatorTron model with a masked-language modeling objective for one epoch, before fine-tuning on NLI4CT. Results from this experiment reveal minor performance gains from the additional training data. \citet{Stanford-2023-nli4ct} @Stanford Mlab uses a combination of back translation, synonym replacement, Random insertions, deletions, and swapping of words on NLI4CT to quadruple the size of the training set. The results presented in Table 2 demonstrate that data augmentation does not inherently result in improved performance, and highlight the importance of selecting suitable tasks, data, or annotations, with respect to the target domain.

\subsection{Biomedical Pre-training}

\paragraph{There is no consistently superior biomedical pre-training strategy.} Models pre-trained on the PubMed Abstract and PubMed Central (PMC) datasets were implemented in 6/21 systems, including the top-performing system \citet{THiFLY-2023-nli4ct} @THiFLY. Additionally, 4/21 systems use models pre-trained on MIMIC III. There is no observable correlation between pre-training data and model performance.

\paragraph{Biomedical pre-training is not sufficient to achieve state-of-the-art performance.} As seen in Table 2 3/5 of the top 5 systems for the entailment task and the evidence selection task do not apply any biomedical pre-training strategies. Furthermore, \citet{Saama-2023-nli4ct} @Saama AI Research demonstrates that large generative models are capable of outperforming the majority-class baseline on the entailment task, even in a zero-shot setting. Additionally,  \citet{Sebis-2023-nli4ct} and \citet{knowcomp-2023-nli4ct} record DeBERTa-v3 \cite{he2021debertav3} significantly outperforming comparatively smaller models pre-trained on biomedical data.

\subsection{Evidence-based NLI}
Many of the discriminative models have a limited input length, often smaller than the CTR premise token length \cite{Clemson-2023-nli4ct}. Therefore extracting a condensed set of evidence facts, prevents the information from being lost by truncation. Even for generative models adapted to receive longer sequences of text, there is still a risk of distractors present in the CTR premise interfering with the inference process, particularly with respect to numerical inference.

\paragraph{Retrieving evidence before inference does not result in better entailment task performance.} Systems that execute the evidence selection task, extract relevant evidence from the premise with respect to the statement, and then use the retrieved evidence for the entailment task (Pre), do not demonstrate significantly higher F1 than models which perform the inference over the entire premise (Post), shown in Table 2. As mentioned previously the cost of excluding relevant information is significant, and systems that perform inference over the entire premise circumvent this cost as they effectively have an evidence extraction Recall of 1.0 at the inference step.

\paragraph{Retrospective evidence retrieval induces confirmation bias.}
11 participants submit to both tasks, and 6 participants opt to classify the entailment, then retrieve evidence from the premise to support the classification. Conversely, 5 participants first extract relevant evidence and then classify the entailment based on selected evidence. There is no significant difference in the results of these clusters for the entailment task, however for the evidence selection task, systems that first collect evidence average +0.045 F1, and +0.07 Precision compared to those that retrospectively select evidence. These clusters report identical average Recall. Therefore, we hypothesize that retrospective systems exhibit confirmation bias, as selected evidence must be relevant to both the statement and the predicted label. The expected effects of reducing the input size by filtering out irrelevant parts of the premise are not evident in the reported results.

\subsection{Limitations}

\paragraph{Joint inference systems may generalize poorly without prior knowledge.} NLI4CT was constructed using a negative-rewriting strategy (Section 5), this results in one contradictory statement and one entailment statement for each CTR premise. \citet{kefah-2023-nli4ct} @JUST-KM and \citet{THiFLY-2023-nli4ct} @THiFLY leverage this feature. These systems perform inference over statement pairs with shared premises and assign the entailment label to the statement with the highest confidence, and then assign the contradiction label to the remaining statement, regardless of confidence. \citet{THiFLY-2023-nli4ct} @THiFLY reports that this process improves entailment task performance by +0.8 F1. The limitations of this approach are that this is heavily reliant on the knowledge that only one statement is entailed, and therefore this approach may generalize poorly where this knowledge is unavailable.

\section{Statistical Artifacts}
 Statistical characteristics such as imbalanced sequence lengths, token distributions, or discriminative conditions that are disproportionately associated with a particular class can superficially inflate model performance \cite{herlihy2021mednli}. \citet{Clemson-2023-nli4ct} @Clemson NLP observes that systems are able to outperform the random baseline on the entailment task using only the statements, this indicates that systems are able to exclusively rely on the presence of superficial statistical patterns within the collection of statements, without learning the underlying rules of the tasks, reporting an F1 of 0.584. This is significantly below the majority baseline (0.66 F1), and as the entailment task is a binary classification task we conclude that the effects of these artifacts on the submitted results are very minimal.

\citet{Clemson-2023-nli4ct} @Clemson NLP identifies minor differences in statement lengths across classes, however, in our analysis we did not find a significant difference. Additionally, we retrieve the 15 most frequently used tokens in the statements for both classes, although we observed some uneven distributions in the training set, these distributions were not present or even inversely correlated in the test set. Therefore, we do not believe either of these characteristics explains the observed results, and claim that NLI4CT is robust to statistical biases.

\section{Conclusion}

This paper presents the systems and results submitted to the SemEval-2023 Task 7 on the NLI4CT dataset. The tasks are challenging, with the majority of submitted systems failing to significantly outperform the majority class baseline on the entailment task. Potentially due to the requirement for sophisticated numerical reasoning, an elevated frequency of biomedical expressions, or the relatively small training set. We observe significantly better performance on the evidence selection task than on the entailment task, and we find that there is no consistent correlation between task performances. The impact of biomedical pre-training is significantly less profound than expected, far outweighed by the effects of increased model size. There is a direct correlation between model size and task performance, with MLMs achieving the highest results in both tasks. There remains room for improvement on both tasks, potentially by exploiting data augmentation to increase training set size, leveraging the zero-shot capabilities of models such as GPT and T5, or through the direct integration of domain knowledge from ontologies. A further error analysis is necessary to evaluate the impact of biomedical pre-training on MLMs, consistency of performance across sections, generalization ability of models trained on NLI4CT, and comparison of performance on numerical versus biomedical instances.

\bibliography{anthology,custom}
\bibliographystyle{acl_natbib}

\appendix

\end{document}